\documentclass[11pt]{article}

\usepackage[letterpaper, margin=1.1in]{geometry}
\usepackage{times}
\usepackage[T1]{fontenc}
\usepackage[utf8]{inputenc}
\usepackage{microtype}
\usepackage{amsmath,amssymb}
\usepackage{graphicx}
\usepackage{booktabs}
\usepackage{multirow}
\usepackage{array}
\usepackage{url}
\usepackage{xcolor}
\usepackage{caption}
\usepackage{float}
\usepackage{longtable}
\usepackage{capt-of}
\usepackage{fancyhdr}
\usepackage[numbers,sort&compress]{natbib}
\usepackage{authblk}
\usepackage{listings}
\usepackage[hidelinks]{hyperref}
\usepackage{xurl}
\usepackage{setspace}

\graphicspath{{figs/}}
\usepackage{placeins}
\makeatletter
\AtBeginDocument{\let\@main@section\section
  \renewcommand{\section}{\FloatBarrier\@main@section}}
\makeatother
\fancypagestyle{plain}{\fancyhf{}\fancyfoot[C]{\thepage}}

\newcommand{\RMSEfifty}{\ensuremath{\mathrm{RMSE}_{50}}}
\newcommand{\gainfifty}{\ensuremath{\mathrm{gain}_{50}}}

\title{\bfseries Do LLMs Know When They Are Wrong About Molecules?\\[2pt]
\large In-Context Property Prediction with Self-Reported Uncertainty}

\author[1]{Matthias Busch\thanks{\raggedright Corresponding authors: \texttt{matthias.busch@tuhh.de}, \texttt{christian.cyron@hereon.de}}}
\author[3]{Marius Tacke}
\author[2,7]{Tamadur Albaraghtheh}
\author[2]{Sviatlana V. Lamaka}
\author[2]{Mikhail L. Zheludkevich}
\author[1,4,5]{Roland C. Aydin}
\author[2,6]{Christian Feiler}
\author[1,3]{Christian J. Cyron\protect\footnotemark[1]}

\affil[1]{Institute for Artificial Intelligence and Simulation in Mechanics, Hamburg University
of Technology, Ei\ss{}endorfer Stra\ss{}e, 21073 Hamburg, Germany}
\affil[2]{Institute of Surface Science, Helmholtz-Zentrum Hereon, Max-Planck-Stra\ss{}e,
21502 Geesthacht, Germany}
\affil[3]{Institute of Material Systems Modeling, Helmholtz-Zentrum Hereon,
Max-Planck-Stra\ss{}e, 21502 Geesthacht, Germany}
\affil[4]{German Research Center for Artificial Intelligence (DFKI),
Stuhlsatzenhausweg, 66123 Saarbr\"ucken, Germany}
\affil[5]{Department of Materials Science and Engineering, Saarland University,
66123 Saarbr\"ucken, Germany}
\affil[6]{Institute for Interface Physics and Engineering, Hamburg University of Technology,
Am Irrgarten, 21073 Hamburg, Germany}
\affil[7]{Institute of Metallic Biomaterials, Helmholtz-Zentrum Hereon, Max-Planck-Stra\ss{}e,
21502 Geesthacht, Germany}

\date{}

\begin{document}
\maketitle
\thispagestyle{plain}

\begin{abstract}
\noindent
Large language models (LLMs) are used in many domains, but molecular property prediction with them is
still little explored. In this work we evaluate the capability of 7 frontier LLMs to predict molecular properties
together with a self-reported uncertainty from in-context examples, and compare them with classical
baselines trained on the same molecules. We use 4 datasets and screen each for contamination of the
training data of the LLMs: three are clean, and the fourth, ESOL, is memorised by most of the models and
serves as a comparison.
On the clean datasets every LLM improves with the context, at 30 examples the most accurate one beats
every baseline. The requested reasoning effort changes accuracy little.
The self-reported uncertainty ranks the prediction errors weakly, with Spearman rank correlations of up
to 0.46, about the level reached by the uncertainties of the baselines.
Combining uncertainty and prediction, on lipophilicity the most certain half of the best LLM is the
most accurate at small context sizes; on BACE and antiviral, and on lipophilicity at large context sizes,
the best LLMs are on par with the best baseline.
On the contaminated ESOL dataset the LLMs show no improvement with the context size, as expected for a
memorised dataset, and their stated uncertainty is smaller on the molecules whose value they recall.
\end{abstract}
\medskip
\noindent Keywords: large language models \textperiodcentered{} in-context learning
\textperiodcentered{} uncertainty quantification \textperiodcentered{} active learning
\textperiodcentered{} selective prediction \textperiodcentered{} data contamination
\textperiodcentered{} molecular property prediction

\section{Introduction}\label{sec:intro}

Large language models (LLMs) predict molecular properties in context from a Simplified Molecular Input
Line Entry System (SMILES) string, and in several studies they reach the accuracy of baseline models
trained on the same small datasets \citep{Jablonka2024,Guo2023,MolecularGPT,Ramos2025,busch2025large}.
Outside chemistry, \citet{Vacareanu2024} found that LLMs given in-context examples solve synthetic
regression tasks about as well as random forests and gradient boosting, and \citet{Requeima2024}
elicited numerical predictive distributions from them. An in-context regressor needs no training, reads
a molecule in the notation chemists use, and can be asked in the same prompt how uncertain its
prediction is.

Uncertainty matters specifically in two scenarios. The first is active learning, where an uncertainty selects the samples to label. Ideally, the samples with the highest prediction error should be taken, but without label, the prediction error is unknown. The second is selective prediction \citep{Geifman2017}, the use of predictions without measuring them: the most certain predictions are
kept, and the error of the kept fraction measures accuracy and ranking together. For both purposes, the
uncertainty must rank the actual prediction errors well; whether the magnitude of the stated uncertainty
also matches the observed errors, its calibration \citep{Kuleshov2018}, is a separate property that this
study does not assess. For classical molecular models ranking and calibration have been compared for
ensembles, sampling and evidential methods
\citep{Hirschfeld2020,Scalia2020}, and stated confidences of language models have been found
over-confident on text and on chemistry questions \citep{Xiong2024,Mirza2025}.
An uncertainty benchmark needs data the models cannot recall, because a recalled value is predicted
with almost no error and nothing requires the stated uncertainty to register that it was recalled
rather than predicted \citep{Busch2026Memorization,Busch2026Blinding}.

In-context regression has a basis outside chemistry: transformers trained on function classes recover
linear and tree-like functions from the examples in their prompt \citep{Garg2022,Akyurek2023}, and
pre-trained LLMs forecast time series \citep{Gruver2023} and classify tabular rows
\citep{Hegselmann2023} from text alone, with an accuracy that grows with the number of examples
\citep{Agarwal2024}. In chemistry, LLMs have served as few-shot predictors of molecular properties
\citep{Jablonka2024,Guo2023,MolecularGPT}, as generators of rules that feed a classical model
\citep{Zheng2025}, and as surrogates for Bayesian optimisation over molecules and catalysts
\citep{Ramos2026,Nguyen2025}; fine-tuned LLM regressors, however, trail dedicated models on materials
data \citep{Jacobs2024}, and LLM surrogates help Bayesian optimisation mainly after adaptation to the
domain \citep{Kristiadi2024}. These studies compare accuracy, mostly at one training-set size; the
learning curve over the number of examples and the overlap of the test molecules with the
pre-training data are rarely reported.

Work on the uncertainty of LLMs has concentrated on classification and question answering
\citep{Geng2024}: models can be taught to state their confidence in words \citep{LinHilton2022}, the
stated confidence is informative \citep{Tian2023} but often too high \citep{Xiong2024,Zhou2023},
prompting alone yields a weaker uncertainty than a short fine-tuning on graded answers
\citep{Kapoor2024}, and the disagreement between repeated samples is an alternative to self-report
\citep{Kuhn2023,Manakul2023}. For numerical prediction the uncertainty has so far been read from token
likelihoods or sampled outputs \citep{Gruver2023,Requeima2024,Ramos2026}, and the one molecular study
we are aware of uses sampling on classification tasks \citep{Chen2024}. The uncertainties of classical
molecular models, in contrast, have been benchmarked across methods \citep{Li2024MUBen}, with the
recommendation to judge them together with the accuracy of the model \citep{Tran2020,Heid2023}, and
they are used for selective prediction \citep{Janet2019} and for choosing the next experiment
\citep{Reker2015,Hie2020}. A like-for-like comparison of the self-reported $\sigma$ of an in-context
regressor with classical uncertainties on the same molecules, judged by these two uses over the number
of examples and on data the models have not seen, has not been made; it is the comparison of this
study.

We evaluate 7 frontier language models on 4 molecular regression datasets: lipophilicity, BACE-1
inhibition, inhibition of the SARS-CoV-2 main protease from a challenge of 2025, and ESOL. The test
molecules of the first three occur in none of seven open pre-training corpora, and a zero-shot screen
finds only insignificant retrieval of their values by any panel model, whereas ESOL is retrieved by most and serves as the control.
Each model predicts 150 test molecules per dataset with a stated standard deviation from nested
contexts of 30 to 1,000 training molecules, in three calls per molecule and, at the small context
sizes, in three independent context draws; eight classical methods are trained on exactly the same
molecules, and six of the models are also run along a ladder of reasoning efforts.
We analyze how accurate the models are over context sizes and reasoning levels, which uncertainty ranks
the prediction errors best, as needed for active learning, how accurate the most certain
predictions are, as needed for selective prediction, and, with the contaminated ESOL dataset as the
comparison, how a memorised dataset behaves on the same measures.

\section{Methods}\label{sec:methods}

\subsection{Datasets and contamination control}\label{sec:data}

We use four molecular regression datasets (Table~\ref{tab:panel}a): lipophilicity, 4,200 octanol/water
distribution coefficients logD at pH 7.4 \citep{MoleculeNet}; BACE, 1,513 inhibitory potencies (pIC50)
against BACE-1 \citep{Subramanian2016}; antiviral, 1,006 inhibitory potencies (pIC50) against the
SARS-CoV-2 main protease released with the ASAP Discovery--Polaris--OpenADMET challenge of 2025
\citep{asap_polaris_openadmet_2025}; and ESOL, 1,128 aqueous solubilities logS \citep{Delaney2004}.
Each dataset is split once into a test set of 150 molecules, drawn from a seeded permutation, and a
training pool. On the three clean datasets a test molecule is eligible only if none of four string
forms, the verbatim SMILES, the RDKit canonical isomeric and non-isomeric SMILES \citep{RDKit} and the
InChIKey, occurs in any of seven open pre-training corpora searched with infini-gram \citep{InfiniGram}:
the training data of OLMo-2 \citep{OLMo}, Dolma \citep{Dolma}, DCLM-baseline \citep{DCLM}, The Pile
\citep{Gao2020Pile}, RedPajama \citep{Weber2024RedPajama} and C4 \citep{Raffel2020C4}; the ESOL test
set is drawn without this filter.
Every pool molecule whose InChIKey connectivity block equals that of a test molecule is removed, so
that no stereoisomer, tautomer or renamed duplicate of a test molecule can enter a context
(Appendix~\ref{app:B}).

Contamination is measured with the corpus search as well as with the
zero-shot screen of \citet{Busch2026Memorization}. In that study, each panel model is asked, with the dataset
named, for the published value of each test molecule to three significant figures, and a model--dataset
cell is flagged when the retention of matching digits exceeds a floor computed from the labels alone
after Benjamini--Hochberg correction \citep{BH1995}. This neither proves contamination nor excludes it, but is an indicator that correlates with the presence of molecules in open pre training corpora \citep{Busch2026Memorization}.

\begin{table}[tbp]
\centering
\small
\setlength{\tabcolsep}{4pt}
\caption{Datasets and model panel. (a) The four datasets: test-set size, deduplicated training pool, largest context size $N$, share of the test molecules found in any of the seven searched pre-training corpora as a SMILES string (in corpus; maximum over the corpora, Fig.~\ref{fig:contamination}b), and the number of panel models the zero-shot screen flags (flagged). (b) The seven language models: whether the endpoint accepts the reasoning effort none (every endpoint accepts minimal to high), the median emitted reasoning tokens per call at the effort low (range over the 24 main-run cells) and how they change along the reasoning ladder (Fig.~\ref{fig:ladder}), and on the ESOL test set the share of the molecules whose published value the model reproduced to three significant figures (hit$_3$; answers given in mol/L converted to logS, Section~\ref{sec:data}) with the outcome of the zero-shot screen and its regime. Endpoints, service tiers and prices: Table~\ref{tab:b_panel}. Endpoints as of 2026-10-08, zero-shot outcomes as of 2026-10-09 (GPT-6 Astra 2026-10-13).}
\label{tab:panel}
\begin{tabular}{@{}lllrrrr@{}}
\multicolumn{7}{@{}l}{(a) Datasets}\\
\toprule
dataset & property & test & pool & largest $N$ & in corpus & flagged \\
\midrule
Lipophilicity \citep{MoleculeNet} & logD at pH 7.4 & 150 & 4,042 & 1,000 & 0.0\,\% & 0 of 7 \\
BACE \citep{Subramanian2016} & BACE-1 pIC50 & 150 & 1,355 & 1,000 & 0.0\,\% & 0 of 7 \\
Antiviral \citep{asap_polaris_openadmet_2025} & SARS-CoV-2 Mpro pIC50 & 150 & 844 & 844 & 0.0\,\% & 0 of 7 \\
ESOL \citep{Delaney2004} & logS (log$_{10}$ mol/L) & 150 & 976 & 976 & 32.4\,\% & 6 of 7 \\
\bottomrule
\end{tabular}
\par\medskip
\begin{tabular}{@{}llrlrl@{}}
\multicolumn{6}{@{}l}{(b) Language models}\\
\toprule
 & & \multicolumn{2}{c}{reasoning tokens per call} & \multicolumn{2}{c}{ESOL zero-shot screen}\\
\cmidrule(lr){3-4}\cmidrule(l){5-6}
model & effort none & at low & along the ladder & hit$_3$ (\%) & outcome \\
\midrule
Haiku 5.5 & accepted & 0 & from medium on & 10.6 & flagged, heavy \\
Sonnet 5.5 & rejected & 0 & none at any effort & 37.2 & flagged, heavy \\
GPT-6.1 Sol & rejected & 74--286 & rise with the effort & 19.5 & flagged, heavy \\
GPT-6 Luna & accepted & 392--581 & rise with the effort & 8.8 & flagged, partial \\
GPT-6 Astra & rejected & 74--298 & no ladder & 18.6 & flagged, heavy \\
Gemini 3.8 Flash & rejected & 0--1,532 & rise with the effort & 12.4 & flagged, heavy \\
GLM 5.3 Flash & rejected & 51--144 & rise only at high & 2.7 & clean \\
\bottomrule
\end{tabular}
\end{table}

\subsection{In-context protocol and model panel}\label{sec:protocol}

The prediction prompt consists of three messages, given verbatim in Appendix~\ref{app:A}: a system
message that assigns the role of a chemist and names the property with its unit, a training block with
one record per molecule, its SMILES string as written in the source file and its label with two
decimals, marked for prompt caching, and a task message with the test SMILES that asks for a prediction
and a 1-sigma uncertainty, defined as the standard deviation of the predictive distribution so that the
measured value falls within prediction $\pm$ uncertainty about 68\,\% of the time, in a fenced JSON
block.
A reply is parsed from its last readable JSON block, and
stated standard deviations below 0.01 are raised to 0.01 for scoring.
The panel consists of seven models served through OpenRouter \citep{OpenRouter}: Claude Haiku 5.5 and
Claude Sonnet 5.5 (Anthropic), GPT-6.1 Sol, GPT-6 Luna and GPT-6 Astra (OpenAI), Gemini 3.8 Flash
(Google) and GLM 5.3 Flash (Z.ai) (Table~\ref{tab:panel}b, Appendix~\ref{app:B}). Each model is pinned
to one provider endpoint with fallback disabled, the flex tier is used where the provider offers one,
every request uses the reasoning effort low and a completion cap of 32,000 tokens, and temperature and
top-p are not sent, so the provider defaults apply; Claude Opus 5.5 is not in the panel because its
endpoint returned a content-filter stop without an answer in every call with 250 or more training
molecules.

\subsection{Context sizes, repeats and context draws}\label{sec:contexts}

The context sizes are $N$ = 30, 60, 120, 250, 500 and the largest available, 1,000 for lipophilicity
and BACE, 844 for antiviral and 976 for ESOL; the in-context set of size $N$ is the first $N$ molecules
of a seeded permutation of the pool, so the sets are nested and identical for every test molecule.
A cell is one combination of dataset, model and context size. Each test molecule is predicted in
$R=3$ calls per cell, and the prediction $\hat y$ and the stated standard deviation $\sigma$ of a
molecule are the means over its parsed repeats. Two further context draws use independent permutations
of the same pool at $N\le120$ with the same test set and settings. Every value in this paper is context
draw 1 at every $N$; at $N\le120$ the range over the three draws is given in brackets, and the draws
are never averaged into a learning curve. Every cell of the main run has an answer for all 150 test
molecules and no call that reached the completion cap.

\subsection{Classical baselines}\label{sec:baselines}

Eight classical methods are trained on exactly the in-context molecules of each cell and predict the
same 150 test molecules (Appendix~\ref{app:C}). TabPFN v3.5 \citep{Hollmann2025,Grinsztajn2026}, a
tabular foundation model trained on synthetic tasks only, receives RDKit two-dimensional descriptors;
its prediction is the mean of its predictive distribution and its $\sigma$ is half the distance
between the 16th and the 84th percentile.
An exact Gaussian process (GP) on Morgan fingerprints uses a Tanimoto kernel \citep{Griffiths2023} with
amplitude and noise chosen by maximising the marginal likelihood \citep{Rasmussen2006}; random forests
\citep{Breiman2001} on the descriptors and on the fingerprints use the spread over the trees as
$\sigma$; deep ensembles \citep{Lakshminarayanan2017} of five Chemprop networks \citep{Heid2024} and of
five multilayer perceptrons use the member spread; a $k$-nearest-neighbour model and a trivial model
that predicts the training mean complete the set.
Every fit receives the labels of its training rows only. TabPFN, both random forests, the GP and the
nearest-neighbour model were run with three seeds in draw 1, and values are averaged over the seeds
within a draw; the paired bootstrap uses seed 0.

\subsection{Metrics and statistics}\label{sec:metrics}

For a cell with test molecules $i=1,\ldots,n$ ($n=150$), $e_i=\hat y_i-y_i$ is the error and
$\sigma_i$ the stated standard deviation of a language model or the $\sigma$ of a baseline. The stated
$\sigma$ is defined operationally, by the prompt of Section~\ref{sec:protocol}, as the standard
deviation of a distribution for the measured value; the $\sigma$ of a baseline is the raw spread
indicator of its method (Section~\ref{sec:baselines}). Between models, $\sigma$ is therefore
compared only through ranks and through the error of the molecules each model ranks as most certain,
not through its values. Accuracy is the root-mean-square error (RMSE). Ranking for measurement selection
is the Spearman rank correlation $\rho(\sigma,|e|)$, where a large $\sigma$ should mark a large error;
ties in $\sigma$ are determined after rounding to 10 decimals.
Selective prediction keeps the $k=\lfloor qn\rfloor$ molecules with the smallest $\sigma$. Stated
standard deviations are coarse, with a median of 30 distinct values among the 150 molecules of a cell,
so several molecules often share the value at the cut.
Let $t$ be the $k$-th smallest $\sigma$, $B=\{i:\sigma_i<t\}$, $T=\{i:\sigma_i=t\}$ with $m=|T|$ and
$r=k-|B|$; if the $r$ places are filled by a uniformly random subset of $T$, the expected mean squared
error of the kept molecules is
\begin{equation}
  \mathrm{MSE}_q=\frac{1}{k}\Bigl(\sum_{i\in B}e_i^2+\frac{r}{m}\sum_{i\in T}e_i^2\Bigr),
  \label{eq:rmseq}
\end{equation}
and $\mathrm{RMSE}_q=\sqrt{\mathrm{MSE}_q}$; the relative error reduction is
$\mathrm{gain}_q=1-\mathrm{RMSE}_q/\mathrm{RMSE}$. \RMSEfifty{} depends on accuracy and ranking
together, while \gainfifty{} isolates the ranking part.
Two sources of uncertainty affect every value beyond the three calls per molecule, which are averaged:
the sampling of the 150 test molecules and the choice of the context molecules. The first is covered by a paired bootstrap over test molecules with 1,000
resamples \citep{Efron1993}, one index matrix per dataset shared by every model and baseline, for the
difference between a language model and TabPFN or the GP; the second by the three context draws of
Section~\ref{sec:contexts}, and at $N\le120$ the bootstrap difference is computed in each draw on the
same resampled molecules and averaged, so that the interval reflects the sampling of test molecules
with the variation between draws averaged out. Intervals are not corrected for multiplicity.

\subsection{Reasoning ladder}\label{sec:ladder}

The reasoning ladder is a sub-experiment with six of the seven models: the efforts none,
minimal, low, medium and high at $N=30$ and at the largest $N$, with $R=3$, the main-run cells serving
as the step low. GPT-6 Astra joined the panel after the ladder, and its ladder would have cost at least
as much as the main run of all seven models.
Because a requested effort does not correspond to the same amount of reasoning across providers, the
results are plotted against the median number of emitted reasoning tokens per call. The effort none
was rejected by four of the six models, and at the effort high the completion cap truncated 15 to
35\,\% of Gemini 3.8 Flash's calls on the clean datasets, so those cells rest on the molecules with a
parsed answer and are censored (Appendix~\ref{app:G}).

\section{Results and Discussion}\label{sec:results}

\subsection{Prediction Accuracy}\label{sec:q1}

\begin{figure}[tbp]
\centering
\includegraphics[width=\textwidth]{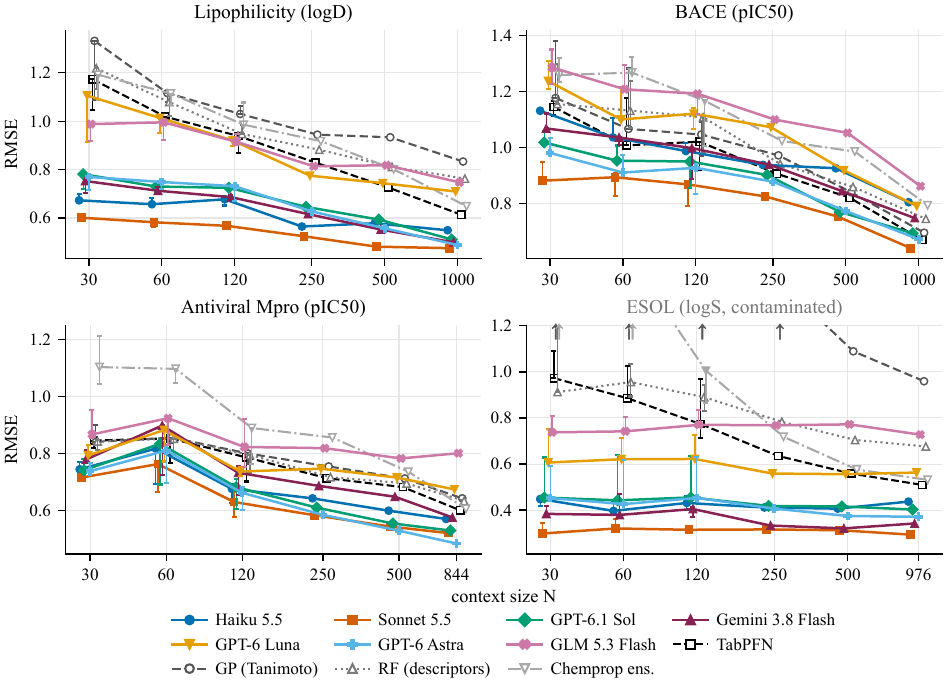}
\caption{Test error against the number of in-context molecules. Root-mean-square error on the 150 test
molecules of each dataset for the seven language models at the effort low and four of the eight
classical baselines trained on the same molecules. Lines follow context draw 1; at $N\le120$ bars span
the range over the three context draws, baselines averaged over their seeds within a draw. ESOL is the
dataset the zero-shot screen flags; its axis is cut, and baselines above the cut are marked by
arrows.}
\label{fig:rmse}
\end{figure}

Fig.~\ref{fig:rmse} shows the RMSE of the seven language models and of four classical baselines against
the number of in-context molecules on the four datasets. On the three clean datasets every model improves
with the context size, the language models by between 8\,\% and 37\,\% per model and dataset from $N=30$
to the largest $N$, whereas on ESOL only the baselines improve and the language models keep an almost
constant RMSE. The curves fall mostly monotonically; the clearest exception is antiviral, where every
language model has a higher RMSE at $N=60$ than at $N=30$ in context draw 1 and none has in draws 2 and 3,
so a single step of this size is a property of the context draw (Appendix~\ref{app:F}).

At $N=30$ the most accurate language model, Sonnet 5.5, has a lower RMSE than every classical baseline
on every dataset in each of the three context draws; on lipophilicity its RMSE is about half that of
TabPFN (0.60 against 1.17, Fig.~\ref{fig:rmse}). The margins differ between the datasets and shrink
with increasing $N$. On ESOL, the dataset the language models have memorised, their learning curves are
flat at a level the baselines do not reach: Sonnet 5.5 has an RMSE of 0.30 at $N=30$ and of 0.29 at
$N=976$, while TabPFN falls from 0.97 to 0.51 over the same context sizes (Section~\ref{sec:contamination}).

How far the memorised knowledge of aqueous solubility carries over to molecules outside the memorised
set is unclear. On lipophilicity the language models improve less with $N$ than the strongest baselines,
by 18 to 37\,\% from $N=30$ to $N=1000$ against 46 to 48\,\% for TabPFN and Chemprop, which indicates
that prior knowledge contributes to their predictions here as well: the models know the property, but
not the values of this dataset (Section~\ref{sec:contamination}). How far this accuracy generalises to
other molecules then depends on how well the models know them. On BACE and antiviral the language
models improve with $N$ about as much as the baselines, apart from a flat start up to $N=120$ for
Sonnet 5.5 and GPT-6 Astra on BACE, which again points to prior knowledge at a lower level.

\begin{figure}[tbp]
\centering
\includegraphics[width=\textwidth]{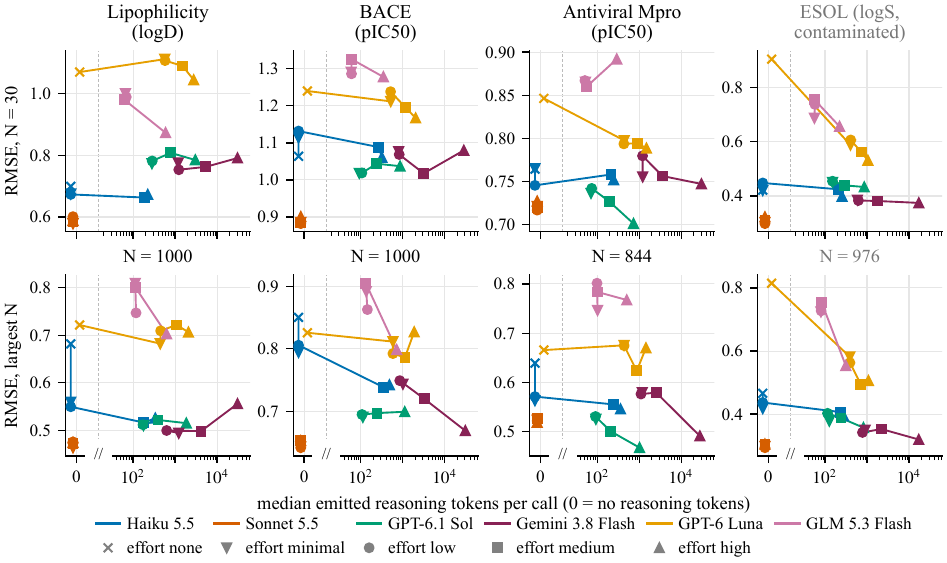}
\caption{Accuracy along the reasoning ladder. Root-mean-square error against the median number of
reasoning tokens a model emitted per call at each requested effort level, for the six ladder models at
30 in-context molecules (top) and at the largest context size (bottom); models that emit no reasoning
tokens are drawn left of the axis break. At the effort high, Gemini 3.8 Flash reached the completion cap
in up to 35\,\% of its calls on the clean datasets, so its points there rest on the remaining molecules.}
\label{fig:ladder}
\end{figure}

Fig.~\ref{fig:ladder} shows the RMSE of the six ladder models against the median number of reasoning
tokens emitted per call at each requested effort level. There is no consistent relation between the
RMSE and the number of reasoning tokens: over the 36 clean cells of the ladder the RMSE at the effort
high differs from that at low by between $-0.12$ and $+0.06$, with no direction shared by the six
models, and the effort none, which only Haiku 5.5 and GPT-6 Luna accept, raises the error of GPT-6 Luna
on ESOL from 0.56 to 0.81.

The number of emitted reasoning tokens varies strongly between models and requested effort levels:
Sonnet 5.5 emits no reasoning tokens at any level, Haiku 5.5 emits them from the effort medium on, and
Gemini 3.8 Flash emits a median of up to 30,717 per call at high.
The cells of Gemini 3.8 Flash at high rest on the 126 to 150 molecules the model finished, a subset
biased towards the molecules it found easy.

\subsection{Uncertainty Quantification}\label{sec:q2}

\begin{figure}[tbp]
\centering
\includegraphics[width=\textwidth]{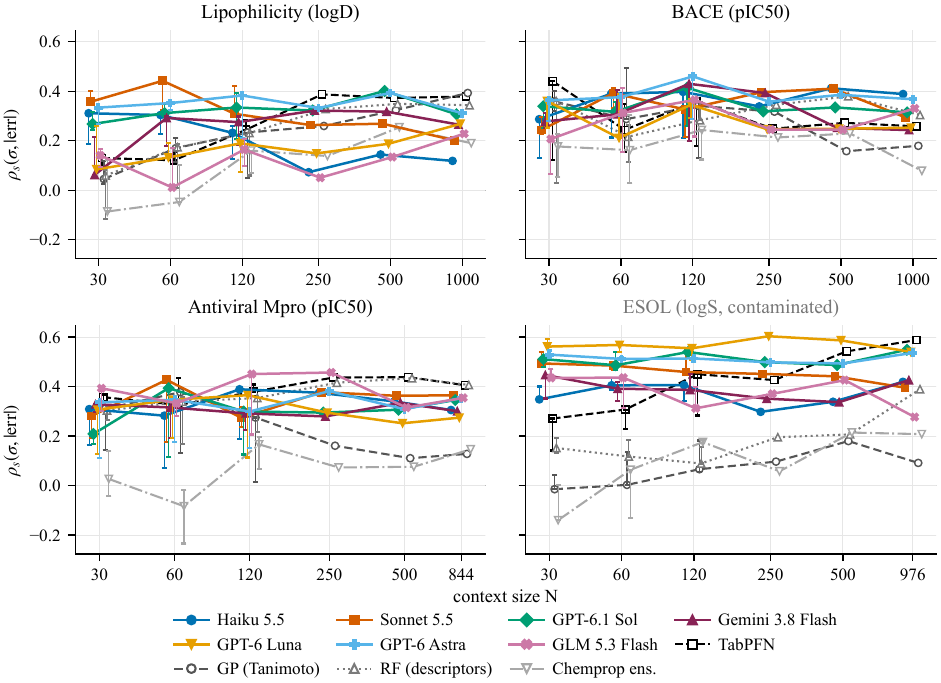}
\caption{Ranking of the errors by the uncertainty, the measure for measurement selection. Spearman rank
correlation between the stated standard deviation and the absolute error on the 150 test molecules
against the number of in-context molecules, for the seven language models at the effort low and four
classical baselines with their own $\sigma$; $\sigma$ rounded to 10 decimals before ranking. Lines and
bars as in Fig.~\ref{fig:rmse}.}
\label{fig:rho}
\end{figure}

Fig.~\ref{fig:rho} shows the Spearman rank correlation between the stated standard deviation and the
absolute error against the context size, for the language models and for the classical baselines with
their own $\sigma$. The stated standard deviation ranks the errors similarly for every model, mostly
between 0.2 and 0.4. On the contaminated ESOL the language models and TabPFN reach higher values of
about 0.4 to 0.6, while the other baselines stay mostly below 0.2. The rank correlation stays at about
the same level across the context sizes; it varies, but with no clear trend of improvement or
deterioration.

For active learning the aim is a model that predicts well after each round, and which molecule to
measure next depends on that model: different models learn differently from the same molecules, and
some bring prior knowledge, as transfer-learned models and language models do. A stated uncertainty is
therefore worth something only together with the accuracy of the model that states it. With few labels
on clean data, up to about 120 molecules, the strongest language models, Sonnet 5.5, GPT-6 Astra,
GPT-6.1 Sol and Haiku 5.5, are probably the better basis for active learning: they predict more accurately than
every classical baseline, and their stated $\sigma$ ranks the errors about as well as a classical
$\sigma$ on the same molecules, better on lipophilicity and about the same on BACE and antiviral
(Section~\ref{sec:q1} and Fig.~\ref{fig:rho}). An actual active-learning loop was not run here, and it
would be needed to confirm this (Section~\ref{sec:limitations}).

\subsection{Selective Prediction}\label{sec:q3}

\begin{figure}[tbp]
\centering
\includegraphics[width=\textwidth]{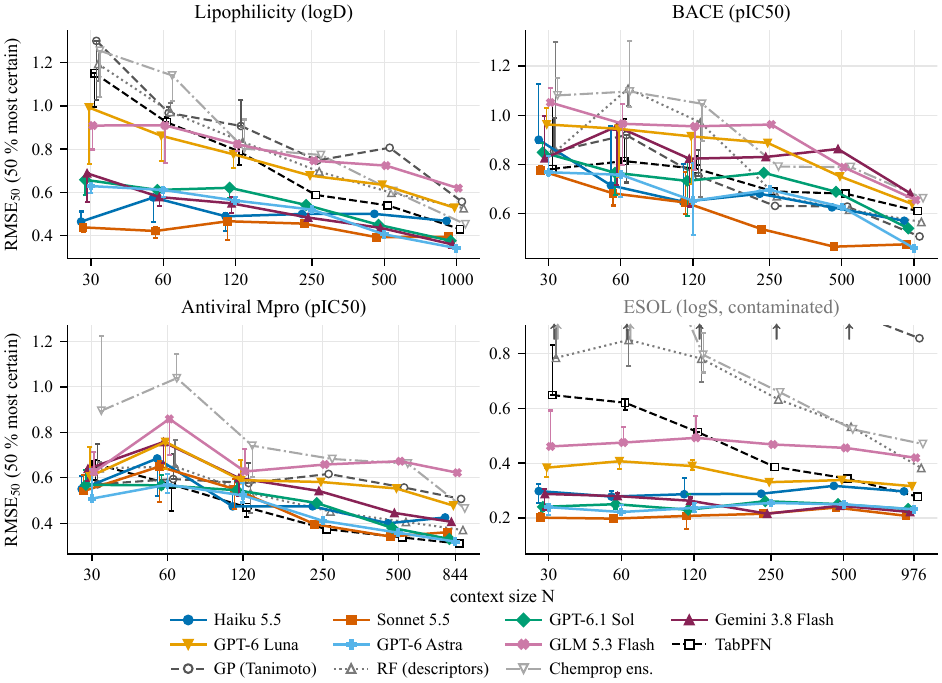}
\caption{Error of the most certain half, the measure for selective prediction. Root-mean-square error
of the 75 test molecules with the smallest stated standard deviation against the number of in-context
molecules, with molecules tied at the cut selected at random (square root of the expected mean squared
error, Eq.~\ref{eq:rmseq}), for the seven language models at the effort low and four classical
baselines. Lines and bars as in Fig.~\ref{fig:rmse}; the ESOL axis is cut, baselines above it are
marked by arrows.}
\label{fig:rmse50}
\end{figure}

Selective prediction replaces measurements by predictions: only the most certain predictions are
used, so it needs a model with a usable uncertainty estimate. \RMSEfifty{} combines the
accuracy and the uncertainty estimate into one number and thus assesses how useful a model is for
this purpose. Fig.~\ref{fig:rmse50} shows the RMSE of the most certain half of the predictions,
\RMSEfifty{}, against the context size. On lipophilicity and on the contaminated ESOL the language
models are ahead of every baseline at most context sizes. As for the RMSE in Section~\ref{sec:q1},
the margins are largest at $N=30$ and shrink with increasing $N$; at $N=1000$ TabPFN reaches the
level of the best language models on lipophilicity. On antiviral the best language models are ahead
of TabPFN at $N=30$ but within 0.06 of the GP, at the same level as TabPFN from $N=60$ on, and
slightly behind it at the largest $N$, except GPT-6 Astra. On BACE Sonnet 5.5 and GPT-6 Astra lie
below TabPFN at every $N$, at $N=30$ by 0.01 in draw 1, and the difference excludes zero only with the
three context draws combined at $N\le120$, not in a single draw (Appendix~\ref{app:F}). The certain
half of the best language model is thus clearly the most accurate only on lipophilicity; on BACE and
antiviral a classical baseline is on par from 30 examples on.

The lead of the language models is smaller in \RMSEfifty{} than in the RMSE of all molecules on BACE and
antiviral, although their $\sigma$ ranks the errors about as well (Section~\ref{sec:q2}): on antiviral
the \gainfifty{} of TabPFN lies between 0.33 and 0.51 from $N=60$ on, against language-model medians of
0.16 to 0.32, whereas on lipophilicity the language models gain at least as much as the baselines at
$N\le120$. The best model for selective prediction therefore depends on the setting. On lipophilicity
the language models remain the best choice, but the \RMSEfifty{} of Sonnet 5.5 and Haiku 5.5 hardly
improves with $N$, so their most certain half rests mostly on prior knowledge, which may not generalise
to other molecules. On BACE and antiviral the best language models are on par with the best baseline or
slightly ahead, and the baselines reach this level at a fraction of the cost (Appendix~\ref{app:H}), so
TabPFN or the GP can be the better option there.

\subsection{Zero-shot correlation, corpus presence and retention}\label{sec:contamination}

\begin{figure}[tbp]
\centering
\includegraphics[width=\textwidth]{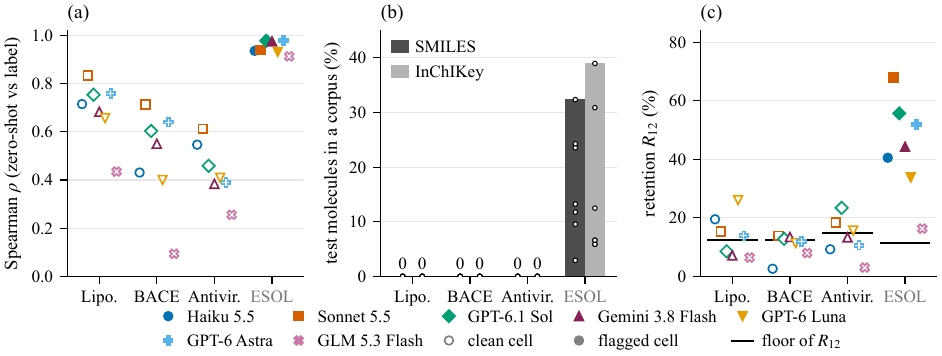}
\caption{Contamination screen of the four test sets. (a) Spearman rank correlation between the labels and
the zero-shot predictions of each language model, asked with the dataset named for the published value of
each test molecule (ESOL: positive answers to molecules with a negative label read as mol/L and converted to logS,
Section~\ref{sec:data}). (b) Share of the test molecules found in the seven searched pre-training corpora as a
SMILES string (verbatim or RDKit canonical) and as InChIKey; bars give the maximum over the corpora, dots
the single corpora (ESOL: the 136 test molecules with a searchable string). (c) Retention $R_{12}$, the
share of the molecules whose zero-shot value matches the label to one significant figure that also match
it to two, against the floor expected from the labels alone (black line). The screen tests $R_{12}$ and
$R_{23}$, the retention from two to three matching figures, against their floors with
Benjamini--Hochberg correction; hit$_3$, the share of the molecules whose published value is reproduced
to three significant figures, is given for ESOL in Table~\ref{tab:panel}b. Filled markers: cell flagged
by the screen; open: clean. Scored are the labels with at least three significant figures, 81
(Lipo., lipophilicity), 143 (BACE), 136 (Antivir., antiviral) and 113 (ESOL); $R_{23}$, hit$_3$, the
floors and the $q$-values of every cell are in the repository.}
\label{fig:contamination}
\end{figure}

Fig.~\ref{fig:contamination} shows the three contamination measurements per model and dataset: the
correlation between the zero-shot predictions and the labels, the share of test molecules found in a
pre-training corpus, and the retention of matching digits against its floor. The three clean datasets
show no significant retrieval and none of their test molecules
occurs in a searched corpus, whereas ESOL is
flagged for six of the seven models (Section~\ref{sec:data}). Without any training molecule the models
already rank the clean test molecules, with Spearman correlations between their zero-shot predictions and
the labels of 0.09 to 0.83, while no clean cell retains matching digits significantly above the floor
computed from the labels alone; the language models thus know the property lipophilicity, not the
values of the dataset. On ESOL all seven models reach correlations of 0.91 to 0.98.

The 63 ESOL test molecules whose value at least one panel model reproduced in the zero-shot screen are
easier for every model than the 50 that no model reproduced, but by different factors: at $N=976$ the
RMSE on the never-recalled molecules is 1.9 to 2.4 times that on the recalled ones for the four models
with the highest hit rate, against 1.3 for TabPFN, which has no access to published values, and for
GLM 5.3 Flash, the one model the screen does not flag. The stated $\sigma$ follows the recall only in
part: it is smaller on the recalled molecules, but by less than the error, so that for Sonnet 5.5 the
root-mean-square error in units of the stated $\sigma$ is 0.42 on the recalled and 1.27 on the
never-recalled molecules (Appendix~\ref{app:D}).

ESOL shows what an uncertainty benchmark measures on a dataset the models have memorised. Accuracy is
constant in $N$ because the prior dominates (Fig.~\ref{fig:rmse}), which also holds for GLM 5.3 Flash,
the model the screen does not flag, so a flat learning curve is not by itself evidence of recall; GLM
5.3 Flash, however, has more than twice the RMSE of Sonnet 5.5 on ESOL. Recall lowers the error on the
recalled molecules more than it lowers the stated $\sigma$, and the spread over the three repeated calls
ranks the errors better than the stated $\sigma$ in 24 of the 42 ESOL cells, where repeats agree on a
recalled value, against 8 of the 126 clean cells (Appendix~\ref{app:E}). An uncertainty benchmark on a memorised dataset therefore
measures in part how confidently a model looks a value up. The protection used here is a property of
the datasets, not of the prompt: blinding reduces retrieval but does not remove it in every case
\citep{Busch2026Memorization,Busch2026Blinding}, and the prompts of this study show real property names
and values.

\subsection{Limitations}\label{sec:limitations}

Each dataset has one test set of 150 molecules, rank correlations over 150 molecules have intervals of
about $\pm0.15$, and the intervals are not corrected for the number of comparisons. The clean datasets
are three, and the context draws cover only $N\le120$; at larger $N$ every value rests on one nested
context chain. The protocol is one prompt with real property names and values, three calls per
molecule and the provider default temperature. The recall flags of ESOL come from one zero-shot call
per model, the corpus search covers open corpora and is a proxy for the closed training data of the
panel models, lipophilicity and BACE originate from ChEMBL and may be present in other forms, and
approximate knowledge of a value cannot be separated from chemical knowledge, so how far the accuracy
generalises to compounds the language models know less about is unclear.
The cells of Gemini 3.8
Flash at the effort high are censored, GPT-6 Astra has no ladder, and the endpoints change without
notice, so every result describes them on the dates given in Section~\ref{sec:data} and
Appendix~\ref{app:H}. The uncertainty was assessed only through its rank correlation with the error and the error of the most certain fraction on fixed test sets; an experiment with an actual active-learning loop, in which the selected molecules are added to the context and the model is queried again, is still outstanding, so the usefulness of the stated uncertainty for active learning is inferred from these two quantities and not measured. The test sets are random splits, so a test molecule can share its scaffold with context molecules; a scaffold split would make the task harder for every model, and probably more so for the classical baselines, which know nothing about a chemical series absent from their training set, than for the language models, which bring prior knowledge of it; on the random splits used here, the error of every model depends only weakly on the similarity of a test molecule to its context (Appendix~\ref{app:E}).

\section{Conclusion}\label{sec:conclusion}

In this study, we evaluated seven frontier LLMs as in-context regressors of molecular properties with a
self-reported uncertainty, on three datasets that a corpus search and a zero-shot screen find
uncontaminated and on the memorised ESOL dataset as a comparison, against eight classical baselines
trained on the same molecules, over context sizes from 30 to 1,000 molecules, three context draws and a
ladder of reasoning efforts.

On the clean datasets every LLM learns from its context, and with few labels the strongest models
predict more accurately than every classical baseline; the margin shrinks with the context size. The
requested reasoning effort changes accuracy little and does not set the amount of reasoning.

The stated uncertainty ranks the errors about as well as a classical uncertainty on the same molecules.
Because an uncertainty is worth something only together with the accuracy of the model that states it,
the strongest LLMs are the better basis for active learning when labels are few, and the distinction
fades with more labels. Whether this holds in an actual active-learning loop remains to be tested.

For selective prediction the most certain half of the best LLM is the most accurate with small
contexts only on lipophilicity, where its predictions are; where an LLM has to learn the property from
the context, as on BACE and antiviral, the best LLMs are only on par with the classical baselines and the baselines reach this level at a fraction of the cost.

On the memorised ESOL dataset the LLMs do not learn with increasing context sizes, as expected when the
models know the dataset, and their uncertainty correlates better with their error than on the clean
datasets, as expected when they know which values they recall and which they have to predict.

\section*{Acknowledgements}
M.B., S.V.L., M.L.Z., C.J.C., R.C.A. and C.F. acknowledge funding by the Deutsche
Forschungsgemeinschaft (DFG; project 535656357). We gratefully acknowledge the support of the
Helmholtz-Gemeinschaft Deutscher Forschungszentren (HGF).

\section*{Funding}
This publication was funded by the Deutsche Forschungsgemeinschaft (DFG; project 535656357):
\url{https://gepris.dfg.de/gepris/projekt/535656357}.

\section*{Author contribution: CRediT}
Matthias Busch: Conceptualization, Methodology, Software, Formal analysis, Investigation, Data
curation, Writing -- Original Draft, Writing -- Review \& Editing, Visualization.
Marius Tacke: Writing -- Review \& Editing.
Tamadur Albaraghtheh: Writing -- Review \& Editing.
Sviatlana V. Lamaka: Writing -- Review \& Editing.
Mikhail L. Zheludkevich: Writing -- Review \& Editing.
Roland C. Aydin: Supervision, Writing -- Review \& Editing.
Christian Feiler: Supervision, Writing -- Review \& Editing.
Christian J. Cyron: Writing -- Review \& Editing.

\section*{Declaration of competing interests}
The authors declare no competing interests.

\section*{Data and code availability}
All code, the prompts, the raw per-call model replies, the splits, the per-cell and per-molecule
tables of every run, the baseline predictions, the contamination-screen replies and flags, the full
bootstrap tables, the per-cell LaTeX tables, and the
scripts that regenerate every figure and table are publicly available at
\url{https://github.com/MatthiasHBusch/LLMsUncertaintyMolecules}. The TabPFN
v3.5 weights are licence-gated and are not redistributed; the library version (tabpfn 9.1.0) and the
checkpoint (\texttt{tabpfn-v3.5-20260909}) are given instead. Endpoint identifiers and prices refer to
OpenRouter as of 2026-10-08. An AI assistant was used during manuscript preparation for writing and
correcting text and for writing the code that produced the tables and figures; all authors have read
the manuscript and take full responsibility for its content.

\bibliographystyle{plainnat}
\bibliography{bibliography}

\clearpage
\appendix
\renewcommand{\thefigure}{A\arabic{figure}}
\renewcommand{\thetable}{A\arabic{table}}
\setcounter{figure}{0}
\setcounter{table}{0}

\noindent This appendix holds the processed material behind the main text: the prompts, the panel and
splits, the baseline settings, the contamination screen, the repeat spread and the context similarity, the selective-prediction
detail, the reasoning ladder and the operational record. Every value per cell, that is per dataset,
model, context size, context draw and reasoning effort, including the per-molecule predictions, the
full paired-bootstrap tables and the per-cell LaTeX tables, is in the repository
named in the data-availability statement and is not reproduced here.

\section{Prompts}\label{app:A}
\lstset{basicstyle=\ttfamily\small, framesep=2pt, aboveskip=2pt, belowskip=2pt}

The three messages for lipophilicity at $N=30$, with the first two of the 30 training records and the
first test molecule (CHEMBL1951022), generated by \texttt{src/icl\_common.py::build\_messages}. The
markers \texttt{**} are part of the prompt text. System message:
\begin{lstlisting}
**Your Role**
You are a professional chemist with expert knowledge in physical chemistry and lipophilicity prediction.
You are tasked with predicting the lipophilicity (octanol/water distribution coefficient logD at pH 7.4) of organic molecules based on their SMILES strings.

**Problem Description**
You will be provided with:
1. A training dataset of molecules with their SMILES and known logD (pH 7.4) values.
2. A test molecule (SMILES) for which you must predict the logD (pH 7.4).

Use your chemical knowledge and the relationships between molecular structure and the logD (pH 7.4) in the training molecules to make an accurate prediction.
\end{lstlisting}
Training block (second message, marked \texttt{cache\_control: ephemeral}):
\begin{lstlisting}
**Training Data:**
30 molecules from the same source as the test molecule, with their SMILES and known logD (pH 7.4) values:

SMILES: Oc1ccc(cc1)C2CNCCc3c(Cl)c(O)c(O)cc23
logD (pH 7.4): 0.87

SMILES: COc1ccc2cc(ccc2c1)[C@H](C)C(=O)O
logD (pH 7.4): 0.13

[... 28 further records in the same format ...]
\end{lstlisting}
Task message:
\begin{lstlisting}
**Prediction Task:**
Predict the octanol/water distribution coefficient logD at pH 7.4 for the following molecule:
- SMILES: CC(C)OC(=O)N1CCC(CC1)Oc2ncnc(Nc3ccc(nc3C)S(=O)(=O)C)c2C

Also report your 1-sigma uncertainty for this prediction, in the same units as the logD (pH 7.4): the standard deviation of your predictive distribution, so that the measured value should fall within prediction +/- uncertainty about 68 % of the time.

**Output (JSON)**
Output the result as a single fenced JSON code block with EXACTLY this structure:
```json
{
  "prediction": ...,
  "uncertainty": ...
}
```
Hard requirements:
- `prediction` is your predicted logD (pH 7.4).
- `uncertainty` is that 1-sigma value (a non-negative number).
- The JSON block MUST be the LAST thing in your response (no text after the closing ```).
\end{lstlisting}
The other datasets substitute the expertise (``medicinal chemistry and BACE-1 inhibitor potency
prediction'', ``medicinal chemistry and SARS-CoV-2 main protease (Mpro) inhibitor potency
prediction'', ``physical chemistry and solubility prediction''), the task (``BACE-1 inhibitory potency
(pIC50)'', ``SARS-CoV-2 main protease (Mpro) inhibitory potency (pIC50)'', ``measured aqueous
solubility logS (log10 of the solubility in mol/L)'') and the label (``BACE-1 pIC50'', ``SARS-CoV-2
Mpro pIC50'', ``measured logS (log10 mol/L)''); everything else is identical.
The parser takes the last fenced JSON block of a reply that can be read, or the outermost
brace-delimited text if there is none, and a reply without both numbers is excluded from the molecule's
repeat means; the parse rate is at least 0.995 in every cell of the main run and of the context draws.
The zero-shot screen uses the prompt of \citet{Busch2026Memorization} unchanged, with the dataset
named; for ESOL it asks an ``expert chemist [who knows] the Delaney (ESOL) dataset very well'' for the
measured aqueous solubility ``as reported in that dataset'' to three significant figures, and the other
datasets substitute the dataset name, property and unit.

\section{Panel, endpoints and splits}\label{app:B}

Table~\ref{tab:b_panel} lists the seven endpoints with their service tiers and prices;
Table~\ref{tab:panel}b gives the reasoning efforts they accept and the reasoning tokens they emit. Every
request names one provider endpoint and disables OpenRouter's fallback to other hosts; the flex tier,
billed at half the standard price, is used for the OpenAI and Google models.
Claude Opus 5.5 answered both calls of an in-context probe with 30 training molecules and returned a
content-filter stop without an answer in both calls with 250 and in all four calls with 1,000 training
molecules.
Further candidates were left out because their reasoning was either switched off or ran to the
completion cap, because the account's privacy setting blocked their only provider, or because of cost;
the probe records are in the repository.

\begin{table}[tbp]
\centering
\small
\setlength{\tabcolsep}{5pt}
\caption{Endpoints and prices. The provider endpoint on OpenRouter to which the requests of each language model are pinned, with fallback to other hosts disabled, its service tier (flex: the provider's flex tier, billed at half the standard price), and the price per million input, cached-input and output tokens at that endpoint. The reasoning efforts an endpoint accepts and the emitted reasoning tokens are in Table~\ref{tab:panel}b. Prices fetched from OpenRouter on 2026-10-08 (\texttt{src/registry/models\_v2.json}, which also gives the OpenRouter ids).}
\label{tab:b_panel}
\begin{tabular}{@{}llllrrr@{}}
\toprule
 & & & & \multicolumn{3}{c}{USD per million tokens}\\
\cmidrule(l){5-7}
model & vendor & provider endpoint & tier & input & cached & output \\
\midrule
Haiku 5.5 & Anthropic & anthropic & standard & 0.1 & 0.01 & 0.5 \\
Sonnet 5.5 & Anthropic & anthropic & standard & 2 & 0.1 & 10 \\
GPT-6.1 Sol & OpenAI & openai/flex & flex & 1 & 0.05 & 5 \\
GPT-6 Luna & OpenAI & openai/flex & flex & 0.05 & 0.005 & 0.25 \\
GPT-6 Astra & OpenAI & openai/flex & flex & 5 & 0.5 & 25 \\
Gemini 3.8 Flash & Google & google-ai-studio/flex & flex & 0.375 & 0.0375 & 1.875 \\
GLM 5.3 Flash & Z.ai & z-ai/fp8 & standard & 0.15 & 0.03 & 0.5 \\
\bottomrule
\end{tabular}
\end{table}

Table~\ref{tab:b_datasets} gives the splits. Every dataset is permuted once with the seed 20260907; on
the clean datasets the test set is the first 150 molecules of the permutation that pass the corpus
filter, which rejected 20 of the first 170 lipophilicity candidates and no BACE or antiviral candidate.
After the test set was fixed, every pool molecule with the InChIKey connectivity block of a test
molecule was removed: enantiomers or stereo-unspecified forms on lipophilicity, stereoisomers and
protonation or tautomer forms on BACE, stereoisomers on antiviral, and on ESOL two compounds that
appear twice under different names with the identical label.
Draws 2 and 3 permute the deduplicated pool of draw 1 with seeds derived from the main seed by SHA-256
and keep the test set; their first 120 molecules share at most 20 molecules with those of another draw,
close to the expectation for independent permutations.
Draw 1 is the split fixed before any model was run; draws 2 and 3 were generated after the main run.

\begin{table}[tbp]
\centering
\small
\setlength{\tabcolsep}{6pt}
\caption{Datasets and splits. Number of molecules; test candidates rejected by the corpus filter (present in a searched corpus) among the positions checked; test-set size; pool molecules removed because their InChIKey connectivity block equals that of a test molecule (dedup.); training pool after removal; and the largest number of molecules shared by the first 120 molecules of two context draws, with the expectation for independent permutations in parentheses. Largest context size, corpus presence and screen outcome: Table~\ref{tab:panel}a and Fig.~\ref{fig:contamination} (\texttt{data/splits/*\_v2*.json}, \texttt{results/meta/context\_draw\_overlap.csv}).}
\label{tab:b_datasets}
\begin{tabular}{@{}lrrrrrr@{}}
\toprule
dataset & molecules & corpus filter & test & dedup. & pool & draw overlap (exp.) \\
\midrule
Lipophilicity & 4,200 & 20 of 170 & 150 & 8 & 4,042 & 5 (3.6) \\
BACE & 1,513 & 0 of 150 & 150 & 8 & 1,355 & 10 (10.6) \\
Antiviral & 1,006 & 0 of 150 & 150 & 12 & 844 & 20 (17.1) \\
ESOL & 1,128 & 0 of 150 & 150 & 2 & 976 & 19 (14.8) \\
\bottomrule
\end{tabular}
\end{table}

\section{Classical baselines}\label{app:C}

Table~\ref{tab:c_baselines} gives the configuration of the eight methods. TabPFN v3.5 (library
tabpfn 9.1.0, checkpoint of 2026-09-09) receives the 217 two-dimensional descriptors of RDKit; per cell, columns
that are constant or more than half missing on the training rows are dropped and missing values are
imputed with the training median. One fit takes 9 to 17\,s on a CPU at $N=30$ and at most 98\,s at
$N=1{,}000$.
The GP maximises the log marginal likelihood of amplitude and noise variance from 14 starts within the
bounds $a\in[10^{-3},10^{2}]\,\mathrm{var}(y)$ and $s^2\in[10^{-4},10^{1}]\,\mathrm{var}(y)$. At $N=30$
the optimum lies on a bound in all 12 cells, four datasets in three draws: in 8 on the lower noise
bound, which interpolates the training labels, and in 4 on the lower amplitude bound, which reduces the
GP to the trivial predictor, so at $N=30$ the GP $\sigma$ is either the training standard deviation or
close to the posterior variance alone.
The ensembles' member spread measures disagreement between members only and is far narrower than the
errors: the root-mean-square error in units of the member spread lies between 3.8 and 26.8 for the Chemprop
ensemble and between 8.9 and 31.8 for the MLP ensemble over the cells of draw 1.
Every fit receives a label vector in which all labels outside the cell's training rows are missing; a
self-check replaces every label outside the training rows of one cell by arbitrary values and confirms
that the predictions and $\sigma$ of every method are bit-identical, and confirms for all stored cells
that the training molecules are the first $N$ of the cell's draw.

\begin{table}[tbp]
\centering
\small
\setlength{\tabcolsep}{4pt}
\caption{Classical baselines. Input features, fixed settings, the definition of the uncertainty $\sigma$ (SD: standard deviation) and the seeds run in context draw 1 / in draws 2 and 3 (values are averaged over the seeds of a draw; the paired bootstrap uses seed 0). Every method is trained on exactly the in-context molecules of a cell and predicts the same 150 test molecules (\texttt{results/baselines/*.meta.json}, \texttt{summary.csv}).}
\label{tab:c_baselines}
\begin{tabular}{@{}>{\raggedright\arraybackslash}p{2.0cm}>{\raggedright\arraybackslash}p{3.8cm}>{\raggedright\arraybackslash}p{3.9cm}>{\raggedright\arraybackslash}p{3.1cm}l@{}}
\toprule
method & features & settings & $\sigma$ & seeds \\
\midrule
TabPFN v3.5 & RDKit 2D descriptors (159--201 after the filter) & v3.5, tabpfn 9.1.0, default estimators, CPU & $(q_{84}-q_{16})/2$ of the predictive distribution & 0,1,2 / 0 \\
GP (Tanimoto) & Morgan r = 2, 2,048 bits & $k=a\,T+s^2\delta$, ML-II from 14 starts & posterior predictive SD incl. noise & 0,1,2 / 0,1,2 \\
RF (descriptors) & RDKit 2D descriptors (as TabPFN) & 200 trees, scikit-learn defaults & SD over the trees & 0,1,2 / 0 \\
RF (Morgan) & Morgan r = 2, 2,048 bits & 200 trees, scikit-learn defaults & SD over the trees & 0,1,2 / 0 \\
Chemprop ensemble & molecular graph (chemprop 2 featurizer) & 5 members, chemprop 2.0.0 defaults, 50 epochs & SD over the members & 0 / 0 \\
MLP ensemble & Morgan r = 2, 2,048 bits & 5 members, 2 $\times$ 256 ReLU, dropout 0.1, Adam 0.001 & SD over the members & 0 / 0 \\
kNN (Tanimoto) & Morgan r = 2, 2,048 bits & $k=5$, Tanimoto-weighted mean & SD of the 5 neighbour labels (floor 0.01) & 0,1,2 / 0,1,2 \\
trivial & -- & training mean & training SD & 0 / 0 \\
\bottomrule
\end{tabular}
\end{table}

\section{Contamination}\label{app:D}

Fig.~\ref{fig:contamination} gives the corpus presence and the zero-shot outcome per dataset and
model; the per-cell values of the screen, the hit rates, retentions, floors and $q$-values, are in the
repository (\nolinkurl{results/contamination/zeroshot/cells.csv}). The flagged ESOL cells fall in the
regime of heavy retrieval for Sonnet 5.5, GPT-6.1 Sol, GPT-6 Astra, Gemini 3.8 Flash and Haiku 5.5 and of
partial retrieval for GPT-6 Luna \citep{Busch2026Memorization}.
The corpus search queried all molecules of the four datasets in seven infini-gram indexes in
four string forms, the ESOL molecules also by name; strings shorter than eight characters were not
queried, which leaves 7,753 of the 7,847 molecules searchable.
Only the training data of OLMo-2 32B Instruct is the actual training set of a model; the other indexes
measure how widely a string is redistributed on the open web, and absence there does not prove absence
from a private crawl. A third of the ESOL test molecules occur as a SMILES string in Dolma.
In the zero-shot screen all replies were parsed and none was truncated.
The unit rule for ESOL (Section~\ref{sec:data}) is a plausibility rule on the sign of the label and an
addition to the protocol of \citet{Busch2026Memorization}, which scores the parsed number as given:
without it Sonnet 5.5 reaches a hit$_3$ of 31.0\,\% instead of 37.2\,\% and Haiku 5.5 8.8\,\% instead of
10.6\,\%, which moves Haiku 5.5 from partial to heavy retrieval, and the zero-shot Spearman correlations
of the two models are 0.27 and 0.51 instead of 0.94 and 0.94.

\begin{figure}[tbp]
\centering
\includegraphics[width=\textwidth]{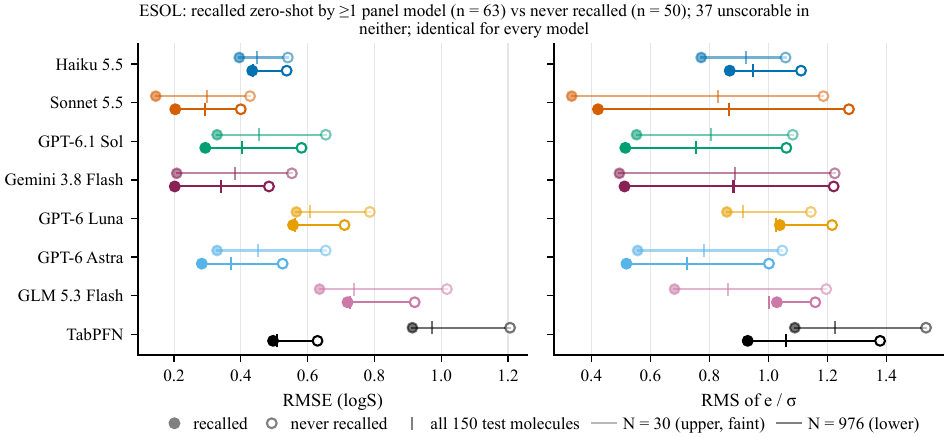}
\caption{Error and error in units of the stated $\sigma$ on recalled and never-recalled ESOL molecules.
Root-mean-square error and root mean square of $e/\sigma$ for the 63 recalled (filled) and the 50
never-recalled (open) test molecules, at 30 (faint) and 976 (solid) in-context molecules; ticks mark all
150 test molecules, and the vertical line marks a root mean square of $e/\sigma$ of 1. TabPFN, which has
no access to published values, is shown for reference, averaged over its seeds.}
\label{fig:esol}
\end{figure}

Fig.~\ref{fig:esol} and Table~\ref{tab:d_esol} split every ESOL cell into the 63 recalled and the 50
never-recalled test molecules; the subsets are the same for every model, because a molecule counts as
recalled when any panel model reproduced it, and the 37 molecules whose label has fewer than three
significant figures belong to neither.
Every model is more accurate on the recalled molecules in every cell, and in every one of the 42
language-model cells the stated $\sigma$ is smaller on the recalled than on the never-recalled molecules,
but by less than the error: the root-mean-square error in units of the stated $\sigma$,
$(\frac1n\sum_i e_i^2/\sigma_i^2)^{1/2}$, is lower on the recalled subset in all 42 cells.
The flat ESOL curves are not explained by recall alone: GLM 5.3 Flash, which the screen does not flag,
has an RMSE of 0.74 at $N=30$ and 0.73 at $N=976$.

\begin{table}[tbp]
\centering
\small
\setlength{\tabcolsep}{4pt}
\caption{ESOL by recall at 976 in-context molecules. Test molecules whose value at least one panel model reproduced to three significant figures in the zero-shot screen (rec., 63 molecules) against those no model reproduced (never, 50); the remaining 37 labels have fewer than three significant figures and are in neither subset. Root-mean-square error with the ratio never / recalled, and root mean square of the error in units of the stated $\sigma$, $(\frac1n\sum_i e_i^2/\sigma_i^2)^{1/2}$ with $\sigma$ floored at 0.01, on all 150 molecules and on the two subsets; language models at effort low, TabPFN v3.5 seed 0, context draw 1 (\texttt{results/icl/calibration\_cells.csv}, columns RMSE and \texttt{s}).}
\label{tab:d_esol}
\begin{tabular}{@{}lrrrrrrrr@{}}
\toprule
 & & \multicolumn{4}{c}{RMSE} & \multicolumn{3}{c}{RMS $e/\sigma$}\\
\cmidrule(lr){3-6}\cmidrule(l){7-9}
model & hit$_3$ \% & all & rec. & never & ratio & all & rec. & never \\
\midrule
Haiku 5.5 & 10.6 & 0.44 & 0.43 & 0.54 & 1.2 & 0.95 & 0.87 & 1.11 \\
Sonnet 5.5 & 37.2 & 0.29 & 0.20 & 0.40 & 2.0 & 0.87 & 0.42 & 1.27 \\
GPT-6.1 Sol & 19.5 & 0.40 & 0.29 & 0.58 & 2.0 & 0.76 & 0.52 & 1.06 \\
GPT-6 Luna & 8.8 & 0.56 & 0.56 & 0.71 & 1.3 & 1.02 & 1.04 & 1.22 \\
GPT-6 Astra & 18.6 & 0.37 & 0.28 & 0.53 & 1.9 & 0.72 & 0.52 & 1.00 \\
Gemini 3.8 Flash & 12.4 & 0.34 & 0.20 & 0.48 & 2.4 & 0.88 & 0.51 & 1.22 \\
GLM 5.3 Flash & 2.7 & 0.73 & 0.72 & 0.92 & 1.3 & 1.00 & 1.03 & 1.16 \\
TabPFN v3.5 & -- & 0.51 & 0.50 & 0.63 & 1.3 & 1.05 & 0.92 & 1.36 \\
\bottomrule
\end{tabular}
\end{table}

\section{Repeat spread and similarity to the context}\label{app:E}

The spread over the three repeated calls of a molecule, the standard deviation of its three
predictions, is a second ranking signal that the repeats provide. It is exactly zero for between 0.7
and 33\,\% of the molecules of a cell.
It ranks the errors better than the stated $\sigma$ in 8 of the 126 clean cells and in 24 of the 42
ESOL cells; for Sonnet 5.5 on ESOL at $N=976$ its rank correlation with the absolute error is 0.65
against 0.40 for the stated $\sigma$, and Sonnet 5.5 returns three identical predictions for 38\,\% of
the recalled and 18\,\% of the never-recalled molecules there.
Combining both signals in the variance of the mixture of the three repeats,
$\sigma_\mathrm{mix}^2=\frac13\sum_r(\sigma_r^2+\hat y_r^2)-\bar{\hat y}^2$ with the prediction
$\hat y_r$ and the stated $\sigma_r$ of repeat $r$ and their mean prediction $\bar{\hat y}$, changes the
rank correlation with the error by a median of 0.00 over the 126 clean cells, upwards in 68 of them and
downwards in the rest, and raises it in all 42 ESOL cells, by a median of 0.03
(\nolinkurl{results/icl/mixture_sigma.csv}).

\begin{figure}[tbp]
\centering
\includegraphics[width=\textwidth]{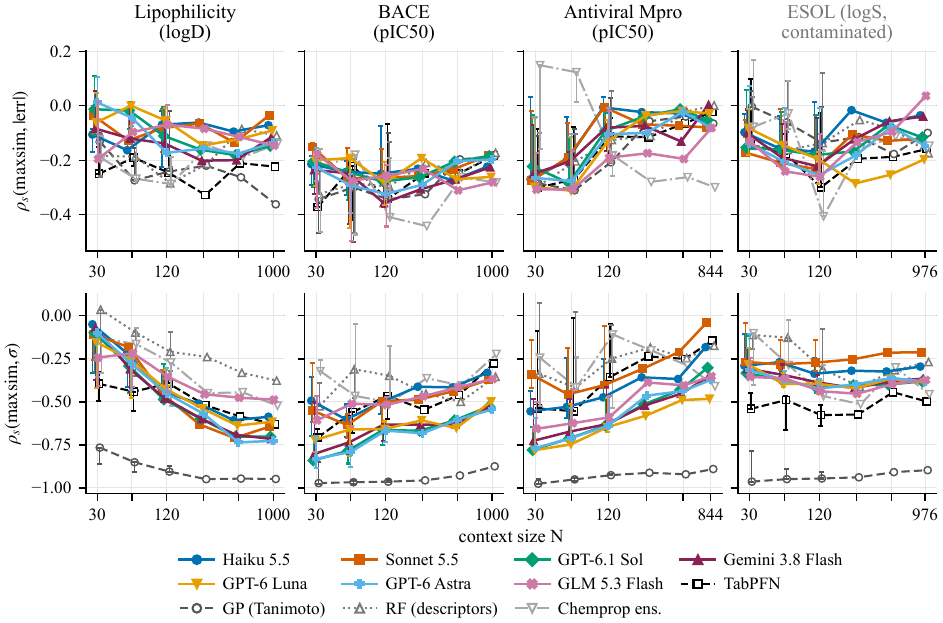}
\caption{Similarity of the test molecules to their context against the error and the stated
uncertainty. Spearman rank correlation between the largest Tanimoto similarity of a test molecule
(Morgan fingerprint, radius 2, 2,048 bits) to any of the $N$ in-context molecules and its absolute
error (top) and its stated standard deviation (bottom), against the number of in-context molecules,
for the seven language models at the effort low and four classical baselines; a negative value means
that molecules with a close context neighbour are predicted more accurately (top) or with a smaller
$\sigma$ (bottom). Lines and bars as in Fig.~\ref{fig:rmse}.}
\label{fig:similarity}
\end{figure}

Fig.~\ref{fig:similarity} relates the error and the stated $\sigma$ to the similarity of a test
molecule to its context, measured as the largest Tanimoto similarity of its Morgan fingerprint to any
of the $N$ in-context molecules. On the clean datasets a close context neighbour predicts a smaller
error only weakly: the rank correlation between this similarity and the absolute error has a median of
$-0.21$ over the language models at $N=30$ and of $-0.12$ at the largest $N$, with a range from $-0.31$
to 0.01 over both sizes, and the four baselines lie between $-0.37$ and 0.15.
The stated $\sigma$ of the language models follows the similarity more closely than their error does,
with a median rank correlation of $-0.56$ at $N=30$ and of $-0.49$ at the largest $N$, more closely
than the $\sigma$ of TabPFN at the largest $N$, $-0.54$ and $-0.28$, and less closely than that of the GP,
$-0.97$ and $-0.89$, whose $\sigma$ is built from the kernel similarities to the context molecules.

\section{Selective prediction and context draws}\label{app:F}

\begin{figure}[tbp]
\centering
\includegraphics[width=\textwidth]{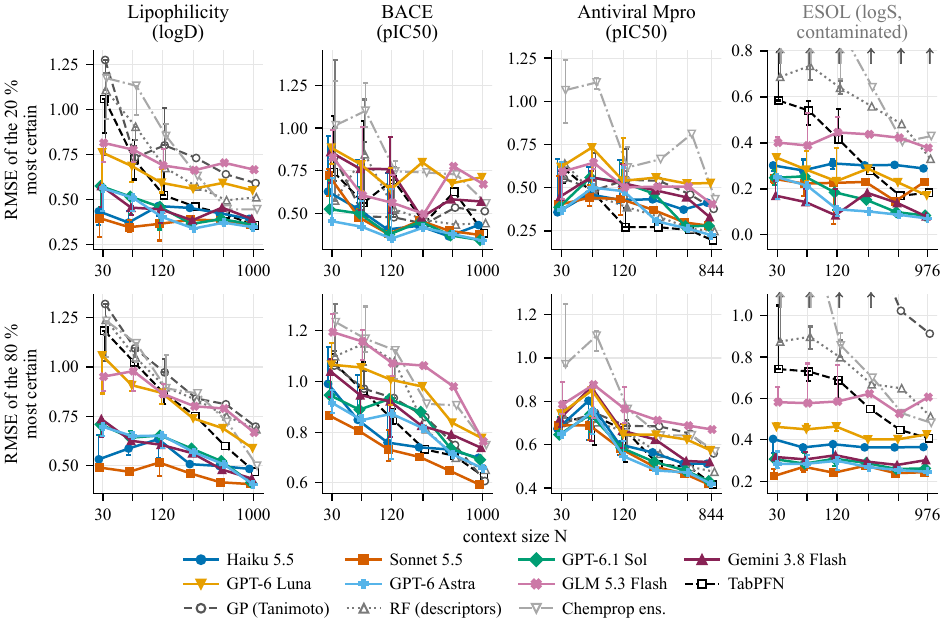}
\caption{Error of the most certain fifth and of the most certain four fifths. Root-mean-square error of
the 20\,\% (top) and the 80\,\% (bottom) of the test molecules with the smallest stated standard
deviation against the number of in-context molecules, with the tie rule of Eq.~\ref{eq:rmseq}, for the
seven language models at the effort low and four classical baselines. Lines and bars as in
Fig.~\ref{fig:rmse}; the ESOL axes are cut, baselines above them are marked by arrows.}
\label{fig:rmseq}
\end{figure}

\begin{figure}[tbp]
\centering
\includegraphics[width=\textwidth]{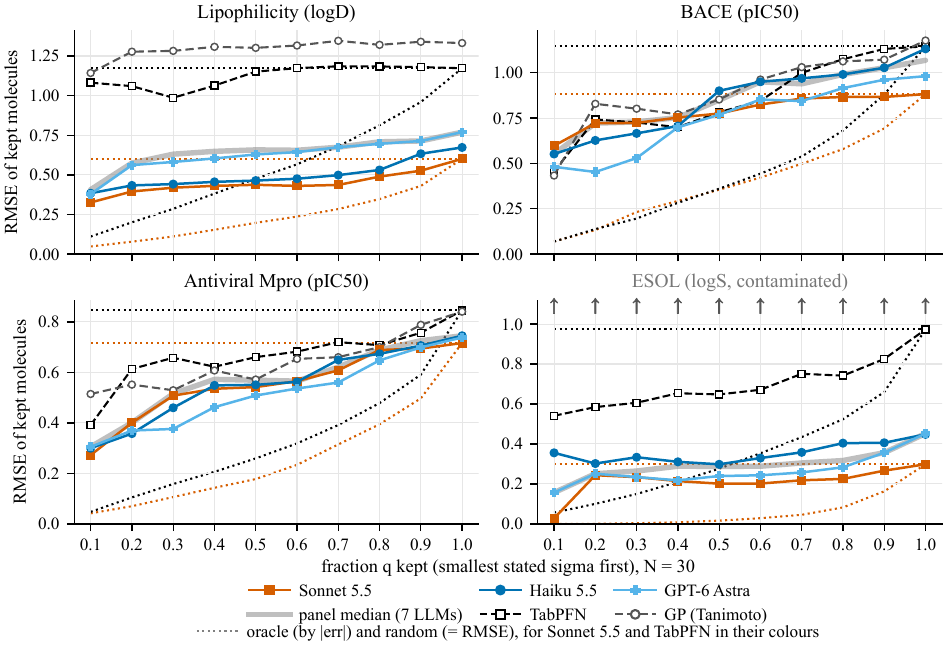}
\caption{Error of the most certain molecules as a function of how many are kept. For each dataset at
$N=30$ (context draw 1), the root-mean-square error of the fraction $q$ of test molecules with the
smallest stated $\sigma$ (tie rule of Eq.~\ref{eq:rmseq}), for Sonnet 5.5, Haiku 5.5, GPT-6 Astra, the
pointwise median of the seven language models (thick grey), TabPFN and the GP (mean over seeds).
Dotted references for Sonnet 5.5 and TabPFN in their colours: the oracle that keeps the molecules with
the smallest absolute error, and random selection. The ESOL axis is cut at 1.12; the GP lies at 1.74 to
1.86 for every $q$ (arrows).}
\label{fig:riskcov}
\end{figure}

\begin{figure}[tbp]
\centering
\includegraphics[width=\textwidth]{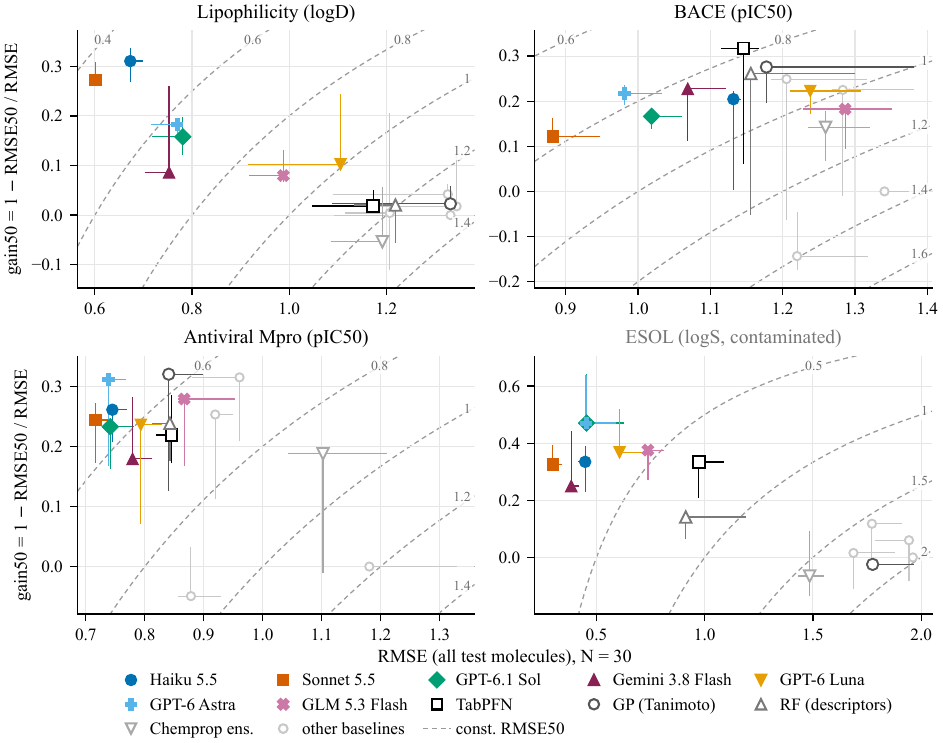}
\caption{Accuracy and selection at 30 in-context molecules. Root-mean-square error of all test
molecules against the relative error reduction obtained by keeping the half with the smallest stated
standard deviation, for the language models (filled) and the classical baselines (open; further
baselines in light grey); markers show context draw 1, bars the range over the three context draws.
Dashed lines connect points of equal error on the selected half.}
\label{fig:joint}
\end{figure}

Fig.~\ref{fig:rmseq} extends Fig.~\ref{fig:rmse50} to the strict cut $q=0.2$ and the loose cut
$q=0.8$. At $q=0.8$ the ordering of the models equals that of the RMSE on all molecules; at $q=0.2$
TabPFN keeps the more accurate fifth on antiviral from $N=120$ on, while on lipophilicity and BACE the
fifth kept by the best language model is as accurate as TabPFN's or better at every $N$
(Section~\ref{sec:q3}). Fig.~\ref{fig:riskcov} shows every cut at $N=30$, and Fig.~\ref{fig:joint}
places accuracy and ranking side by side at $N=30$: the most accurate language models stand to the
left of every baseline on all three clean datasets, but above the baselines only on lipophilicity.
Because the stated standard deviations take few distinct values, the cut of \RMSEfifty{} splits a tied
group in 286 of the 336 language-model cells of the main run and the context draws; the exact range of
\RMSEfifty{} over all tie orders has a median width of 0.03 and is at most 0.23 in a draw-1 cell, so
the tie rule of Eq.~\eqref{eq:rmseq} is needed for a deterministic value.

\begin{table}[tbp]
\centering
\small
\setlength{\tabcolsep}{9pt}
\caption{Paired bootstrap at 30 in-context molecules, language model minus TabPFN v3.5. Difference with the percentile 95\,\% interval from 1,000 resamples of the 150 test molecules, computed in each of the three context draws on the same resampled molecules and averaged (TabPFN seed 0). $\dagger$ / $\ddagger$: the interval excludes 0 in favour of the language model / of TabPFN. The intervals at $N$ = 60, 120 and the largest $N$, and against the GP, are in the repository (\texttt{results/icl/bootstrap\_cells.csv}).}
\label{tab:f_bootstrap}
\begin{tabular}{@{}llll@{}}
\toprule
model & Lipophilicity & BACE & Antiviral \\
\midrule
\multicolumn{4}{@{}l}{$\Delta$RMSE}\\
Haiku 5.5 & $-$0.42 [$-$0.53, $-$0.29]$^{\dagger}$ & $-$0.01 [$-$0.15, $+$0.15] & $-$0.07 [$-$0.16, $+$0.01] \\
Sonnet 5.5 & $-$0.50 [$-$0.60, $-$0.41]$^{\dagger}$ & $-$0.23 [$-$0.34, $-$0.13]$^{\dagger}$ & $-$0.10 [$-$0.18, $-$0.03]$^{\dagger}$ \\
GPT-6.1 Sol & $-$0.36 [$-$0.47, $-$0.25]$^{\dagger}$ & $-$0.11 [$-$0.19, $-$0.03]$^{\dagger}$ & $-$0.07 [$-$0.16, $+$0.01] \\
GPT-6 Luna & $-$0.11 [$-$0.23, $+$0.03] & $+$0.11 [$-$0.00, $+$0.25] & $-$0.02 [$-$0.13, $+$0.06] \\
GPT-6 Astra & $-$0.37 [$-$0.47, $-$0.26]$^{\dagger}$ & $-$0.14 [$-$0.21, $-$0.07]$^{\dagger}$ & $-$0.07 [$-$0.16, $+$0.01] \\
Gemini 3.8 Flash & $-$0.37 [$-$0.45, $-$0.28]$^{\dagger}$ & $-$0.05 [$-$0.13, $+$0.03] & $-$0.03 [$-$0.13, $+$0.05] \\
GLM 5.3 Flash & $-$0.15 [$-$0.24, $-$0.08]$^{\dagger}$ & $+$0.14 [$+$0.06, $+$0.23]$^{\ddagger}$ & $+$0.06 [$-$0.05, $+$0.17] \\
\addlinespace
\multicolumn{4}{@{}l}{$\Delta\rho(\sigma,|e|)$}\\
Haiku 5.5 & $+$0.14 [$-$0.01, $+$0.30] & $-$0.06 [$-$0.16, $+$0.04] & $-$0.12 [$-$0.21, $-$0.02]$^{\ddagger}$ \\
Sonnet 5.5 & $+$0.26 [$+$0.09, $+$0.43]$^{\dagger}$ & $-$0.02 [$-$0.13, $+$0.09] & $-$0.12 [$-$0.25, $-$0.00]$^{\ddagger}$ \\
GPT-6.1 Sol & $+$0.15 [$-$0.02, $+$0.31] & $+$0.03 [$-$0.07, $+$0.13] & $-$0.16 [$-$0.26, $-$0.05]$^{\ddagger}$ \\
GPT-6 Luna & $+$0.03 [$-$0.15, $+$0.18] & $+$0.03 [$-$0.06, $+$0.13] & $-$0.11 [$-$0.22, $+$0.00] \\
GPT-6 Astra & $+$0.17 [$+$0.00, $+$0.33]$^{\dagger}$ & $+$0.04 [$-$0.07, $+$0.14] & $-$0.12 [$-$0.23, $-$0.01]$^{\ddagger}$ \\
Gemini 3.8 Flash & $+$0.03 [$-$0.13, $+$0.19] & $+$0.00 [$-$0.11, $+$0.11] & $-$0.04 [$-$0.14, $+$0.07] \\
GLM 5.3 Flash & $+$0.01 [$-$0.12, $+$0.14] & $-$0.11 [$-$0.22, $+$0.00] & $+$0.01 [$-$0.11, $+$0.12] \\
\addlinespace
\multicolumn{4}{@{}l}{$\Delta$RMSE$_{50}$}\\
Haiku 5.5 & $-$0.58 [$-$0.69, $-$0.47]$^{\dagger}$ & $+$0.02 [$-$0.10, $+$0.11] & $-$0.06 [$-$0.13, $+$0.02] \\
Sonnet 5.5 & $-$0.63 [$-$0.77, $-$0.51]$^{\dagger}$ & $-$0.17 [$-$0.30, $-$0.05]$^{\dagger}$ & $-$0.08 [$-$0.16, $+$0.02] \\
GPT-6.1 Sol & $-$0.43 [$-$0.55, $-$0.30]$^{\dagger}$ & $-$0.07 [$-$0.19, $+$0.02] & $-$0.03 [$-$0.10, $+$0.05] \\
GPT-6 Luna & $-$0.23 [$-$0.36, $-$0.09]$^{\dagger}$ & $+$0.05 [$-$0.07, $+$0.16] & $+$0.02 [$-$0.08, $+$0.11] \\
GPT-6 Astra & $-$0.45 [$-$0.57, $-$0.32]$^{\dagger}$ & $-$0.16 [$-$0.27, $-$0.04]$^{\dagger}$ & $-$0.05 [$-$0.13, $+$0.03] \\
Gemini 3.8 Flash & $-$0.46 [$-$0.56, $-$0.33]$^{\dagger}$ & $-$0.05 [$-$0.20, $+$0.08] & $-$0.03 [$-$0.12, $+$0.07] \\
GLM 5.3 Flash & $-$0.19 [$-$0.31, $-$0.07]$^{\dagger}$ & $+$0.14 [$+$0.01, $+$0.28]$^{\ddagger}$ & $+$0.03 [$-$0.05, $+$0.11] \\
\addlinespace
\multicolumn{4}{@{}l}{$\Delta$gain$_{50}$}\\
Haiku 5.5 & $+$0.26 [$+$0.10, $+$0.39]$^{\dagger}$ & $-$0.03 [$-$0.12, $+$0.08] & $+$0.01 [$-$0.12, $+$0.10] \\
Sonnet 5.5 & $+$0.25 [$+$0.11, $+$0.39]$^{\dagger}$ & $-$0.02 [$-$0.11, $+$0.07] & $+$0.01 [$-$0.14, $+$0.12] \\
GPT-6.1 Sol & $+$0.12 [$-$0.03, $+$0.26] & $-$0.01 [$-$0.10, $+$0.09] & $-$0.03 [$-$0.15, $+$0.07] \\
GPT-6 Luna & $+$0.13 [$-$0.01, $+$0.26] & $+$0.03 [$-$0.07, $+$0.14] & $-$0.05 [$-$0.16, $+$0.08] \\
GPT-6 Astra & $+$0.13 [$-$0.01, $+$0.28] & $+$0.05 [$-$0.07, $+$0.15] & $-$0.00 [$-$0.14, $+$0.10] \\
Gemini 3.8 Flash & $+$0.14 [$+$0.01, $+$0.28]$^{\dagger}$ & $+$0.01 [$-$0.10, $+$0.13] & $+$0.01 [$-$0.12, $+$0.12] \\
GLM 5.3 Flash & $+$0.05 [$-$0.06, $+$0.16] & $-$0.02 [$-$0.11, $+$0.07] & $+$0.02 [$-$0.10, $+$0.13] \\
\bottomrule
\end{tabular}
\end{table}

\begin{figure}[tbp]
\centering
\includegraphics[width=\textwidth]{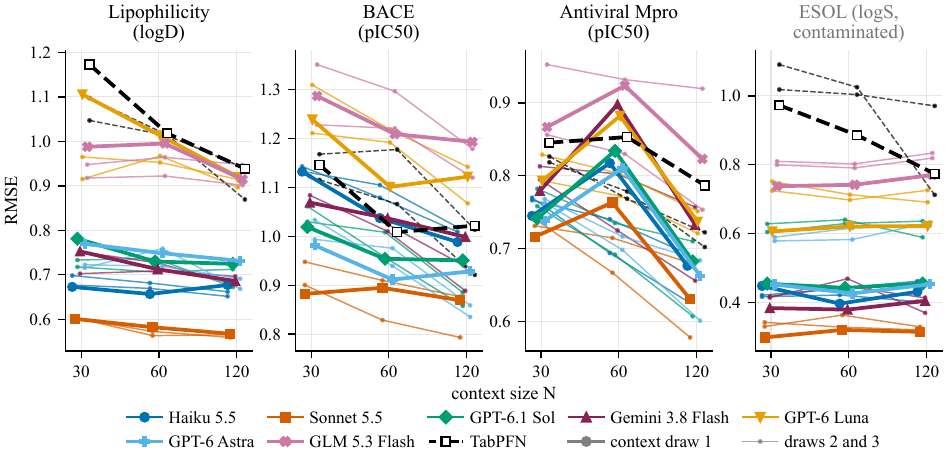}
\caption{Learning curves of the three context draws. Test RMSE at $N$ = 30, 60 and 120 for the seven
language models and TabPFN (averaged over its seeds within a draw), one line per context draw: draw 1,
the nested chain of all other figures, bold, and draws 2 and 3 thin; models slightly offset in $N$.}
\label{fig:draws}
\end{figure}

Fig.~\ref{fig:draws} shows the learning curves of the three context draws at $N\le120$. The median
standard deviation of the RMSE over the three draws, taken over the models of a dataset and context
size, is lower for the language models than for the baselines in 10 of the 12 combinations of dataset
and $N$, which fits predictions that lean on a prior more than on the particular training molecules.
Table~\ref{tab:f_bootstrap} gives the paired-bootstrap differences to TabPFN at $N=30$ that the main
text quotes; the bootstrap interval of a single rank correlation over 150 molecules has a median
half-width of 0.15, and the \gainfifty{} of one language model at $N=30$ differs by up to 0.22
between context draws, so a selection difference of that size from a single draw is not interpretable.

\section{Reasoning ladder}\label{app:G}

Fig.~\ref{fig:ladder} shows the emitted reasoning tokens and the accuracy per model, effort and context
size; the values of every ladder cell, including the change of the rank correlation from
the effort low to high, are in the repository (\nolinkurl{results/icl/ladder_cells.csv}). Five of the
seven panel endpoints reject the effort none with HTTP status 400, so the ladder has four steps for
Sonnet 5.5, GPT-6.1 Sol, Gemini 3.8 Flash and GLM 5.3 Flash and five for Haiku 5.5 and GPT-6 Luna, 160
cells beyond the main run.
Outside Gemini 3.8 Flash at high, at most two of the 450 calls of a cell were truncated.
Zero reasoning tokens do not mean that a model does not reason: the Anthropic models write visible
reasoning while reporting no reasoning tokens, and Gemini 3.8 Flash reports no reasoning tokens in
18\,\% of its main-run calls and writes its reasoning into the visible answer instead.
The ladder steps other than low cost USD 3.01 (none), 71.62 (minimal), 78.65 (medium) and 241.68
(high), of which Gemini 3.8 Flash at high accounts for USD 171.70.
For GPT-6 Luna and GPT-6.1 Sol the steps other than low cost 4.8 and 6.0 times their step low; applied
to the step-low cells of GPT-6 Astra, a ladder would have cost USD 315 to 393, at least as much as the
whole main run of seven models (USD 311.02).

\section{Operational findings}\label{app:H}

\begin{figure}[tbp]
\centering
\includegraphics[width=\textwidth]{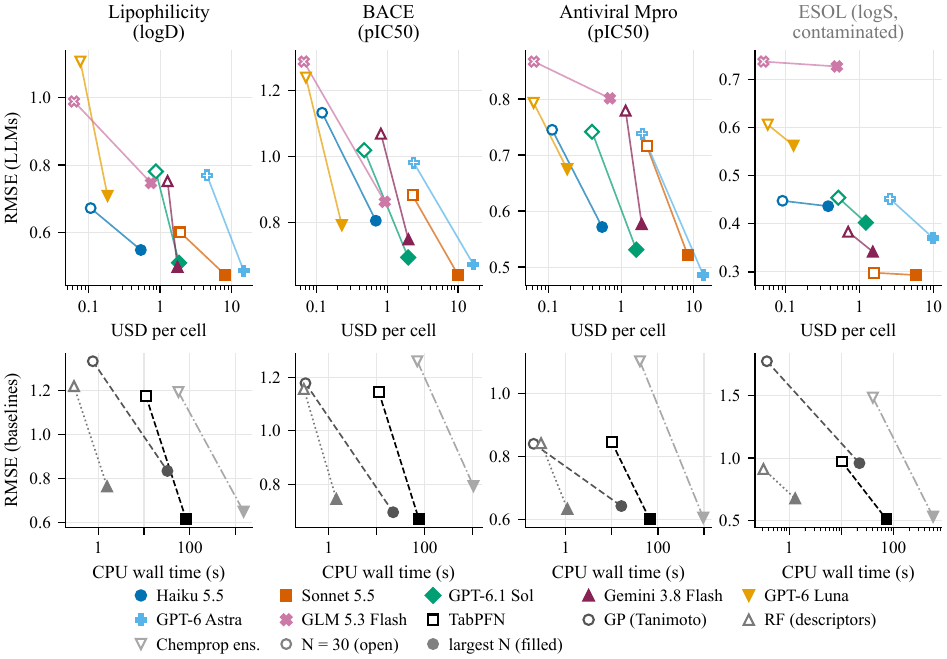}
\caption{Cost and error of each model at the smallest and the largest context. Top: OpenRouter cost in
USD of one cell (150 test molecules, $R=3$ calls, effort low, context draw 1; log scale) against the
test RMSE, open marker $N=30$, filled marker largest $N$. Bottom: four classical baselines, CPU wall
time per cell in seconds (mean over seeds; log scale) against the RMSE; seconds and dollars are not on
one axis.}
\label{fig:cost}
\end{figure}

The calls, the billed cost and the truncated and declined calls of every cell are in the repository
(\nolinkurl{results/icl/cells.csv}, \nolinkurl{results/contamination/zeroshot/cells.csv}).
Endpoints decline requests in ways that shape a panel: besides the Claude Opus 5.5 refusals
(Appendix~\ref{app:B}), the OpenAI endpoints declined single prompts with a reference to their content
policy, mostly antiviral prompts; the missing calls were requested again with the identical prompt, and
the cells with fewer than 450 stored calls after these gap-fill runs are all OpenAI cells, with at least
445 calls and every test molecule answered.
The prompt cache serves most of the context: in the main-run cells with $N\ge120$ the cached share of
the prompt tokens is between 0.91 and 1.00.
The 450 replies of a main-run cell state between 7 and 37 distinct standard deviations, which is why
the selective metrics need a rule for ties.
The context draws contain 1,613 duplicated records, written when two processes worked on the same
cells; they were billed and are excluded from the analysis.
Endpoint lists and prices were fetched from OpenRouter on 2026-10-08, the zero-shot screen ran on
2026-10-09 (GPT-6 Astra on 2026-10-13), the main run between 2026-10-09 and 2026-10-13, the reasoning
ladder on 2026-10-12 and 2026-10-13 and the context draws on 2026-10-13; hosted endpoints change
without notice, so every measurement describes the endpoint as served on its date.
Requests were sent by a Julia runner that stores every reply with its token counts, cached tokens,
finish reason, serving provider and billed cost; the analysis uses Python with numpy, scipy,
scikit-learn \citep{Pedregosa2011} and RDKit, and the baselines tabpfn 9.1.0, chemprop 2.0.0 and
PyTorch.
Fig.~\ref{fig:cost} sets the cost of a cell against its error.

\FloatBarrier

\end{document}